\documentclass[conference, compsoc]{IEEEtran}
\IEEEoverridecommandlockouts
\UseRawInputEncoding
\usepackage{amsmath,amssymb,amsfonts}
\usepackage{algorithmic}
\usepackage{graphicx}
\usepackage{textcomp}
\usepackage{xcolor}
\usepackage{caption}
\usepackage{subcaption}

\ifCLASSOPTIONcompsoc
  \usepackage[nocompress]{cite}
\else
  \usepackage{cite}
\fi

\ifCLASSINFOpdf
\else
\fi

\def\BibTeX{{\rm B\kern-.05em{\sc i\kern-.025em b}\kern-.08em
    T\kern-.1667em\lower.7ex\hbox{E}\kern-.125emX}}
\begin{document}

\title{Leveraging Wastewater Monitoring for COVID-19 Forecasting in the US: a Deep Learning study
}

\author{\IEEEauthorblockN{1\textsuperscript{st} Mehrdad Fazli}
\IEEEauthorblockA{\textit{School of Data Science} \\
\textit{University of Virginia}\\
Charlottesville, VA \\
mf4yc@virginia.edu}
\and
\IEEEauthorblockN{2\textsuperscript{nd} Heman Shakeri}
\IEEEauthorblockA{\textit{School of Data Science} \\
\textit{University of Virginia}\\
Charlottesville, VA \\
hs9hd@virginia.edu}
}

\maketitle

\begin{abstract}

The outburst of COVID-19 in late 2019 was the start of a health crisis that shook the world and took millions of lives in the ensuing years. Many governments and health officials failed to arrest the rapid circulation of infection in their communities. The long incubation period and the large proportion of asymptomatic cases made COVID-19 particularly elusive to track. However, wastewater monitoring soon became a promising data source in addition to conventional indicators such as confirmed daily cases, hospitalizations, and deaths. Despite the consensus on the effectiveness of wastewater viral load data, there is a lack of methodological approaches that leverage viral load to improve COVID-19 forecasting. This paper proposes using deep learning to automatically discover the relationship between daily confirmed cases and viral load data. 
We trained one Deep Temporal Convolutional Networks (DeepTCN) and one Temporal Fusion Transformer (TFT) model to build a global forecasting model. 
We supplement the daily confirmed cases with viral loads and other socio-economic factors as covariates to the models. Our results suggest that TFT outperforms DeepTCN and learns a better association between viral load and daily cases. We demonstrated that equipping the models with the viral load improves their forecasting performance significantly. Moreover, viral load is shown to be the second most predictive input, following the containment and health index. Our results reveal the feasibility of training a location-agnostic deep-learning model to capture the dynamics of infection diffusion when wastewater viral load data is provided.

\end{abstract}


\section{Introduction}
SARS-COV-2 (COVID-19) is a highly contagious infection that has taken the world by storm and impacted millions of lives worldwide since its emergence in late 2019. In September 2021, the mortality rates of COVID-19 surpassed that of the 1918 influenza pandemic (Spanish flu), one of the deadliest pandemics in modern history.
One major challenge that COVID-19 has posed to governments and officials across the globe is predicting its progression in the communities and developing appropriate safety measures to avoid loss of lives and diminish its financial damage. In addition, striking a balance between saving lives by placing restrictions, lockdowns, and curfews; and protecting the economy from massive recession requires accurate prediction of the infection incidences by localities and allocating resources accordingly.

One of the hallmarks of the COVID-19 pandemic is its considerable proportion of asymptomatic or undocumented cases~\cite{nishiura2020rate,pei2021burden,li2020substantial}. Another characteristic of COVID-19 that makes it particularly difficult to curb is its long incubation period, estimated to be nearly six days~\cite{zhang2020evolving,alene2021serial,elias2021incubation}. Presymptomatic individuals can be infectious and unknowingly spread the virus while their infection remains
dormant~\cite{jones_estimating_2021}. In light of the aforementioned findings, researchers have resorted to SARS-COV-2 viral RNA concentration (viral load) in wastewater as an early indicator of incoming peaks of cases~\cite{shah2022wastewater,galani2022sars,zhu2021early, scott2021targeted}. All COVID-19 cases, including presymptomatic and asymptomatic cases, excrete SARS-COV-2 RNA regardless of severity of illness~\cite{jones_estimating_2021}. Thus, viral load captures a critical information source missed in conventional indicators such as clinically confirmed cases, hospitalizations, and deaths. With the growing evidence on the efficiency of wastewater surveillance, the Centers for Disease Control and Prevention (CDC) launched its National Wastewater Surveillance System (NWSS) in September 2020 in an attempt to coordinate wastewater surveillance across the US and help local officials to implement the surveillance system in more localities across the US~\cite{CDC-NWSS}.

Common choices for modeling infectious diseases are compartmental models (e.g., SEIR), autoregressive models (e.g., ARIMA), and machine learning models~\cite{kamalov2022deep}. Compartmental models are the most common models for infectious diseases and have been applied to COVID-19 since the beginning of the outbreak~\cite{ihme2021modeling, he2020seir}. They offer a mathematical formulation for disease evolution in patients and the transmission mechanism. Although these models are highly transparent and interpretable, their simplifying assumptions on population homogeneity and parameter estimations challenges are restricting their use. 
On the other hand, machine learning models do not require significant domain knowledge. They are generalizable, domain-agnostic, easily transferable to different geographies, and, nowadays, easy to train. However, they are less interpretable than compartmental models and require a large amount of data for their training stage.

Some attempts have been made to incorporate viral load data into the compartmental modeling of COVID-19~\cite{fazli2021wastewater,nourbakhsh2022wastewater}. However, incorporating viral load into a compartmental model proves to be challenging. Some of the challenges are determining the appropriate time lag between viral load and incidence data~\cite{kaplan2021aligning}, determining the proper way of incorporating viral load into the mathematical model, taking into account the dilution of SARS-COV-2 RNA and its degradation from the source to the sampling site~\cite{zhu2021early, shah2022wastewater}, the complexity of shedding profile of users and its dependence on the infection severity~\cite{zhu2021early,polo2020making}, etc.
In this paper, we resort to Deep Learning (DL) to eliminate the need for hand-crafting a mathematical formulation for infection transmission and including the necessary covariates at the cost of losing some transparency. To the best of our knowledge, this is the first study that uses viral load data as another input to a DL-based forecasting model. Our focus in this study is to make accurate predictions for the number of COVID-19 infections in the near future. Thanks to health organizations worldwide and the rapid circulation of data, we now have access to detailed COVID-19 data worldwide. That means we have hundreds of time series of daily confirmed cases in different granularity levels. Therefore, a sufficiently large deep learning model can be trained to learn the cross-series patterns. 
DL-based forecasting models such as Long-Short-term memory (LSTM) have been widely applied to forecasting COVID-19~\cite{kamalov2022deep}. We use two probabilistic, multi-step forecasting DL-based models that allow for covariate time series; Deep Temporal Convolutional Networks (DeepTCN) and Temporal Fusion Transformer (TFT).

The contributions of this paper are threefold:

\begin{itemize}
    \item Exploring the feasibility of supplementing COVID-19 incidence data with viral load data to enhance forecasting accuracy.
    \item Exploring the role of socio-economic factors in forecasting the infections.
    \item Training a global probabilistic multi-horizon forecasting model to predict COVID-19 cases in the near future.
\end{itemize}



\subsection{Related Works}
Since the beginning of the COVID-19 pandemic in late 2019, researchers have turned to deep learning models to forecast the propagation of COVID-19 infection~\cite{kamalov2022deep, shoeibi2020automated}. 
This can be attributed to the capacity of these models to learn complex and nonlinear patterns. Some of the models that have been studied in the literature are multi-layer perceptron~\cite{kafieh2021covid, marzouk2021deep}, LSTM~\cite{devaraj2021forecasting, zeroual2020deep, fitra2022deep, shahid2020predictions, marzouk2021deep, elsheikh2021deep}, gated recurrent units (GRU)~\cite{zeroual2020deep, shahid2020predictions}, temporal convolutional networks (TCN)~\cite{wang2022model}, variational autoencoder (VAE)~\cite{zeroual2020deep} and attention-based networks (e.g., TFT)~\cite{jin2021inter, er2021county, fitra2022deep, basu2022covid, zhou2022interpretable}.

Zeroual et al.~\cite{zeroual2020deep} conducted a comparative study on five DL models, simple RNN, LSTM, bidirectional LSTM, GRU, and VAE. They applied the models to confirmed cases and recovered cases data from Italy, France, Spain, China, USA, and Australia. They made predictions for 17 days ahead and concluded that VAE outperforms other models convincingly. Devaraj et al.~\cite{devaraj2021forecasting} also performed a comparative study on autoregressive integrated moving average (ARIMA), LSTM, stacked LSTM, and PROPHET models. They investigated the practicality of using deep learning to forecast cumulative cases globally and achieved an error rate of less than 2\% with stacked LSTM. They also compiled a list of previous publications on COVID-19 forecasting using deep learning. Kamalov et al.~\cite{kamalov2022deep} published one of the few review papers on COVID-19 DL-based forecasting models. They selected 53 papers published between April 2020 and February 2022. They organized the literature by classifying papers based on their modeling approaches and establishing a model-based taxonomy.

Transformers~\cite{vaswani2017attention} are attention-based neural network architectures for sequence-to-sequence modeling, often outperforming recurrent-based neural networks~\cite{vaswaniattention2017,karita2019comparative}. With Transformer's success, researchers started to adapt the model and tune it to other types of machine learning problems.

Er et al.~\cite{er2021county} implemented an attention-based neural network with an encoder-decoder structure named COURAGE. However, unlike the original transformer model, COURAGE uses a linear decoder to make predictions. COURAGE predicts the number of deaths per county in the US for two weeks ahead. COURAGE aggregates county-level data and uses mixup \cite{zhang2017mixup} as a data augmentation method to make more accurate predictions on a state level. They compared the mean absolute error (MAE) of their model to that of the baselines and showed that COURAGE beats most of the baselines. 
Zhou et al.~\cite{zhou2022interpretable} proposed Interpretable Temporal Attention Network (ITANet), a transformer-based model, to forecast the number of COVID-19 cases and infer the importance of government interventions. ITANet is a multi-task learning model whose primary objective is to minimize the loss of the target series (confirmed cases). However, it also has a covariate forecasting network (CNF) that predicts the values of some unknown covariates for the forecast horizon. Unknown covariates are the ones that have to be measured, and their future values are unknown. They also used the Oxford COVID-19 government response tracker~\cite{hale2021global} and temperature and air quality indices as other covariates for their model. They examined their model on state-level data from Illinois, California, and Texas. Their model outperformed the baseline models, including TFT and transformer model, by a considerable margin. They also analyzed the covariate importance and ranked the government intervention indices based on their importance in the three states. 

Jin et al.~\cite{jin2021inter} developed an attention-based model called Attention Crossing Time Series (ACTS) focusing on learning the cross-series recurring patterns. Their method is based on creating embeddings for different time series segments to find similar trends in other time series. They also feed their model with several covariates such as total population, population density, and available hospital beds. They showed that ACTS could outperform the leading models in forecasting COVID-19 incidence (confirmed cases, death, hospitalization) in most cases. The forecasts are made for a four-week horizon and on a state-level granularity. Fitra et al.~\cite{fitra2022deep} applied a Deep Transformer~\cite{wu2020deep} model to Indonesia's COVID-19 data. Deep Transformer adjusts the original transformer model by placing the layer normalization before the LSTM positional encoder. They also compared the Deep Transformer model with RNN and LSTM. Their results suggest that Deep Transformer with AdaMax optimizer outperforms RNN and LSTM models. Basu and Sen~\cite{basu2022covid} applied the TFT model to predict country-wise COVID-19 incidences. They used COVID-19 data from 174 countries and their socio-economic factors, such as social capital, population density, life expectancy, and healthcare access quality, to forecast the cumulative cases for 270 days ahead. 


\section{Methodology}
We intend to leverage the predictive power of neural networks and their capability to learn complicated, recurring, and long-range patterns. Peccia et al. showed that viral load contains valuable information that precedes other indicators (e.g., confirmed daily cases) by several days \cite{peccia2020measurement}. Therefore, it can guide the model to make more accurate predictions. To test this hypothesis, given that viral load is an unknown covariate, we need to choose models that allow for unknown covariates. We are also interested in a global model that can be trained on multiple geographical locations' COVID-19 data and is general enough to make reasonably accurate projections for unseen locations. We argue that such a model has learned the intrinsic dynamics of infection. We select the models based on the paradigms of probabilistic forecasting to quantify the uncertainty and multi-horizon forecasting. Given these criteria, we chose Temporal Fusion Transformers~\cite{lim2021temporal} and DeepTCN~\cite{chen_probabilistic_2020} as our candidate models. Firstly, we introduce each model, then present their results and compare their performance.


\subsection{Forecasting time series}
Assume that we have $U$ related time series, each associated with a location. We aim to learn a general model on all spatially related time series. This general model learns the dynamics of the target time series and its associations with the covariates. 

Given a look-back window of size $k$, from time $t$, we assume a forecast horizon of size $\tau$. Hence the past window is $[t-k, t-k+1, ...,t]$ and the future window is $[t+1, t+2, ...,t+\tau]$. Also, for the duration of the forecast horizon, we suppose that $\textbf{z}^{(u)}_i$ and $\textbf{x}^{(u)}_i$ are unknown and known covariates, respectively. Then the forecasting problem can be summarized as~\cite{lim2021temporal}:
\begin{align}
    \hat{y}^{(u)}(q, t, \tau) = f_q(\tau, y^{(u)}_{t-k:t}, \textbf{z}^{(u)}_{t-k:t}, \textbf{x}^{(u)}_{t-k:t+\tau})
\end{align}
where $y^{(u)}_t$ is the actual $u^{th}$ series and $\hat{y}^{(u)}(q, t, \tau)$ is the predicted $q^{th}$ quantile of $\tau$-step-ahead forecast of $u^{th}$ target series. We note that TCN and TFT produce $\tau$-step-ahead forecast simultaneously as opposed to iterative methods like RNN-based models.

We quantify the population behavior driven by the enforced restrictions using the Oxford Covid-19 government response tracker (OxCGRT)~\cite{hale2021global}. Furthermore, we assume the indices are known for the forecast horizon since any change in governmental policies and restrictions guidelines is usually announced days or sometimes weeks in advance. 
On the other hand, the viral load is a real-time emission from epidemics and has to be measured. Thus viral loads are instead measurements of the dynamical states and not covariates. From a practical point of view, however, viral load can guide the model as a covariate, and its past measurements are available when making predictions. Thus, abusing the terms and considering it as another covariate seems justifiable.

\subsection{Deep Temporal Convolutional Networks}
\begin{figure}
     \centering
     \begin{subfigure}[b]{0.75\columnwidth}
         \centering
         \includegraphics[width=\textwidth]{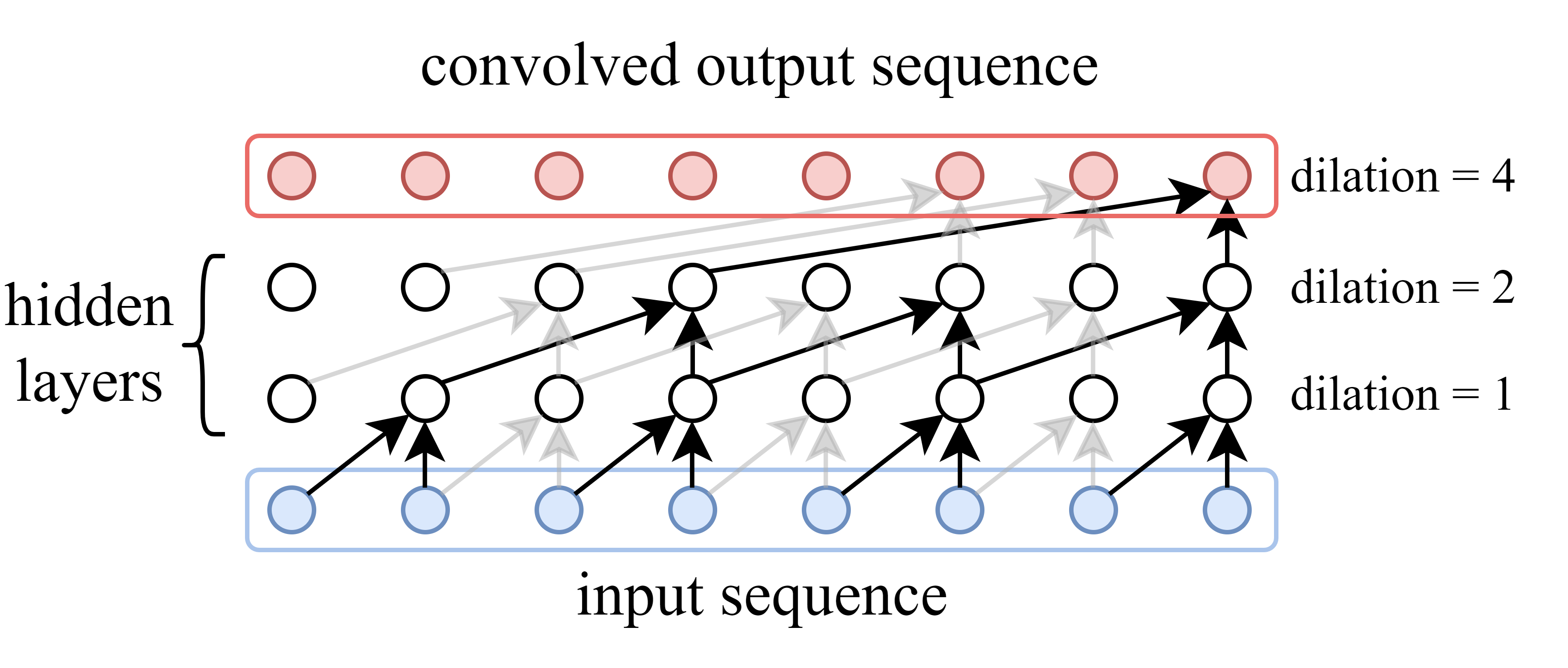}
         \caption{}
         \label{fig:dilated-causal-conv}
     \end{subfigure}
     \hfill
     \begin{subfigure}[b]{0.85\columnwidth}
         \centering
         \includegraphics[width=\textwidth]{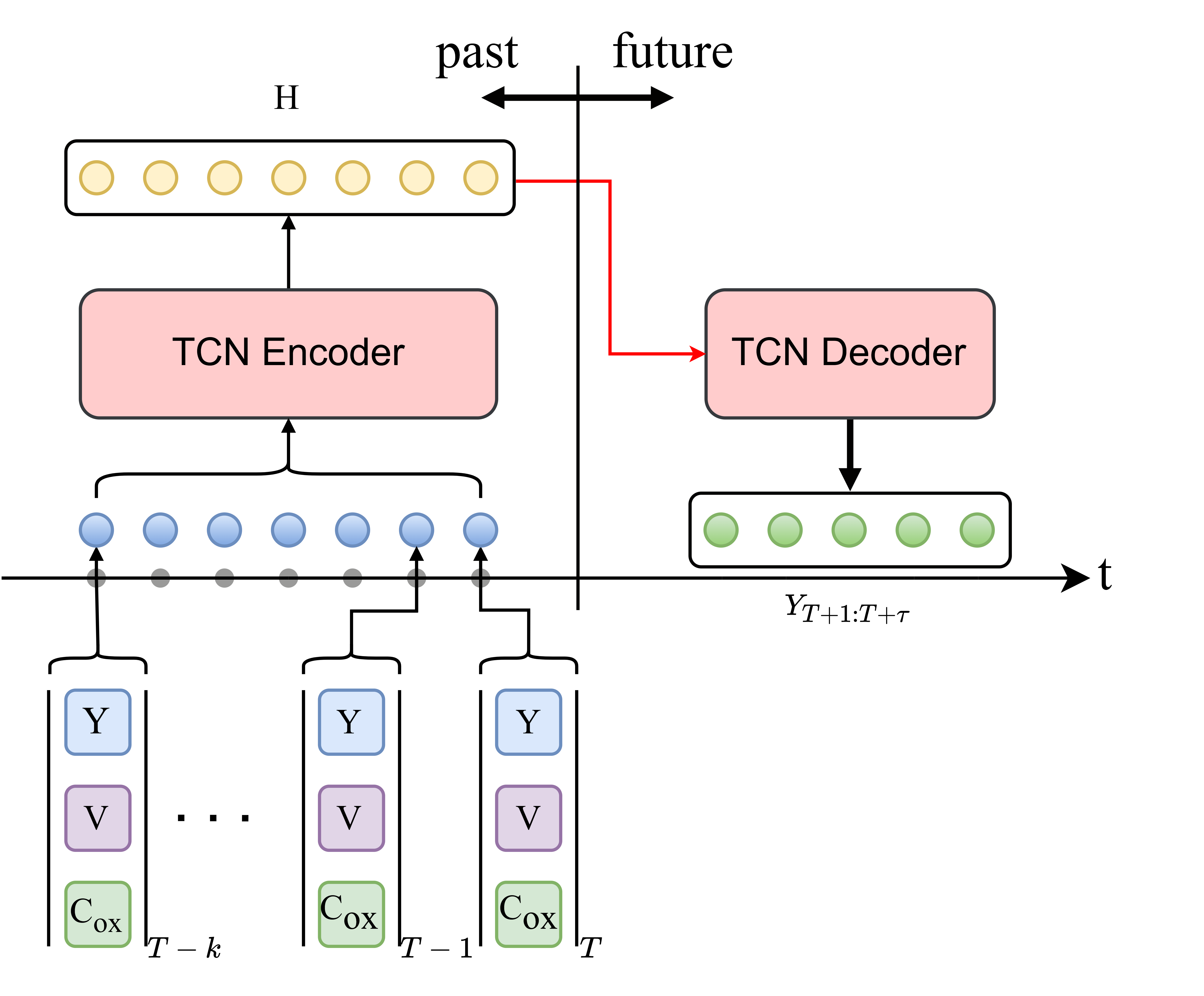}
         \caption{}
         \label{fig:TCN-block}
     \end{subfigure}
        \caption{(a) A dilated causal convolutional network with two hidden layers and dilation factors 1, 2, and 4. (b) Schematic of an encoder-decoder DeepTCN with the inputs and outputs shown. Y, V, and C denote daily cases, viral load, and covariates.}
        \label{fig:TCN}
\end{figure}

Bai et al.~\cite{bai_empirical_2018} compared TCN and different types of RNN (simple RNN, LSTM, GRU) and found that for most of the sequence modeling tasks, TCN outperforms the RNN models. Additionally, they conclude that TCN converges considerably faster than RNN models owing to its non-autoregressive nature.
TCN uses a 1-D fully convolutional network~\cite{long2015fully} to map the input sequence into the output sequence. It also takes advantage of causal convolutional layers to avoid leakage of data from the future into the past. However, stacking convolutional layers in this manner can only increase the look-back window linearly with the number of layers. Therefore, to achieve a long receptive field, one has to stack many convolutional layers, drastically increasing the model's computational burden and making it difficult to train. Van den Oord et al~\cite{oord_wavenet_2016} suggested dilated causal convolutions, which only applies convolution on nodes that are multiplicative of the dilation factor in the previous layer instead of consecutive nodes. For instance, with the dilation factor being one, we restore the regular causal convolutions, while with the dilation factor being two, the model looks at every other node. Fig. \ref{fig:dilated-causal-conv} depicts a dilated causal convolutional network with three different dilation factors for each layer.

A more recent study by Chen et al.~\cite{chen_probabilistic_2020} proposed a modified version of TCN (DeepTCN) for probabilistic forecasting on multiple time series. They suggested an encoder-decoder structure that takes covariates in addition to the target series as inputs. The encoder consists of two dilated causal convolutional networks with batch normalization and ReLU activation function after each. It also has a residual connection inspired by RESNET \cite{he2016deep}. The decoder resembles the encoder, with convolutional networks replaced with dense layers and applied to known covariates. The residual connection combines the output of the encoder with transformed covariates.
DeepTCN makes probabilistic forecasts either by predicting the parameters of a predefined future distribution or by predicting the quantiles of the target value. These modifications added much more flexibility and value to TCN, particularly for the task of time series forecasting. Therefore, we use DeepTCN in our study for its added value while retaining all the characteristics of TCN, such as the long receptive field.



\subsection{Temporal Fusion Transformer}

\begin{figure}
    \centering
    \includegraphics[width=\columnwidth]{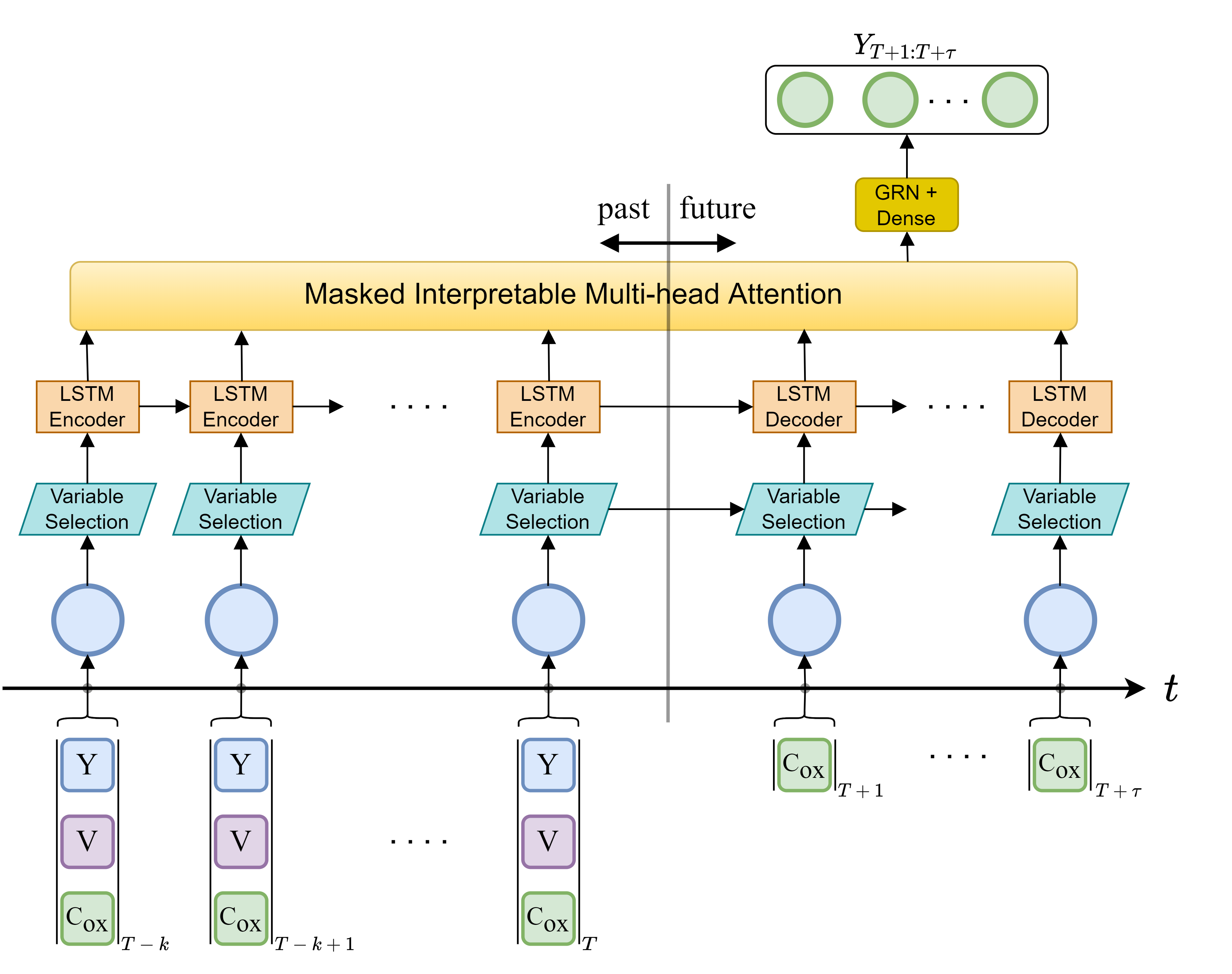}
    \caption{Temporal Fusion Transformer. Y, V, and C denote daily cases, viral load, and covariates.}
    \label{fig:TFT}
\end{figure}

Another successful deep learning architecture for time series forecasting is Temporal Fusion Transformer (TFT)~\cite{lim2021temporal}. Similar to the Transformer model, TFT has an encoder-decoder structure. TFT (Fig. \ref{fig:TFT}) is designed to accept all types of inputs in addition to target series, such as static covariates, known covariates, and unknown covariates. All inputs are processed through a Gated Residual Network (GRN), which is a gating unit with a skip connection to allow the model to skip nonlinear transformation for any input when necessary. Moreover, all inputs (historical data, static and dynamic covariates) are further processed through a variable selection unit to find their weighted average to be fed into the LSTM encoder/decoder. These weights allow us to quantify the importance of each input in the trained model. LSTM acts as a positional encoding unit in the Transformer model. All known inputs are processed through the encoder, while the decoder takes the output of the encoder and known covariates. Similar to the Transformer model, temporal patterns are learned through a multi-head attention mechanism with the difference that the key weights are not head-specific. That enables us to quantify which inputs the model is attending to (relative importance of the past information in forecasting the future).

\subsection{Data}
We use the Biobot Analytics dataset \cite{BiobotData} on national wastewater monitoring. It contains weekly measures of SARS-COV-2 viral genome copies for more than 70 counties across the US. It also includes the number of cases extracted from USA Facts\footnote{https://usafacts.org/issues/coronavirus/} scaled for 100k people. Daily cases were computed by subtracting cumulative counts for two consecutive days, and then a 7-day rolling average was applied.
To enrich the models with socioeconomic and behavior change information, we used the OxCGRT dataset collected at Oxford University~\cite{hale2021global}. This dataset contains several indices, quantifying each state's government response, public health restrictions, and economic support. In particular, we selected four indices, \textit{overall government response index}, \textit{containment and health index}, \textit{stringency index}, and \textit{economic support index}. For simplicity, we call these four covariates Oxford covariates. These indices serve as known covariates whose future values are known during the forecast horizon.

To prepare the data for modeling, we performed several preprocessing steps. As the viral load data is sparser, we focused on the time window in which we have viral load data. Furthermore, to avoid data loss, we changed the granularity of the viral load data from weekly to daily by interpolating the days between two consecutive recordings. We also cut off the tail of data for counties with ragged daily cases due to insufficient recordings. Additionally, we scaled the viral load, case count, and oxford covariates values to be between 0 and 1. Finally, we created four date-time covariates, namely year, month, week of the year, and day of the week, to help the model extract seasonality in the data.

\begin{figure}
     \centering
     \begin{subfigure}[b]{0.95\columnwidth}
         \centering
         \includegraphics[width=\textwidth]{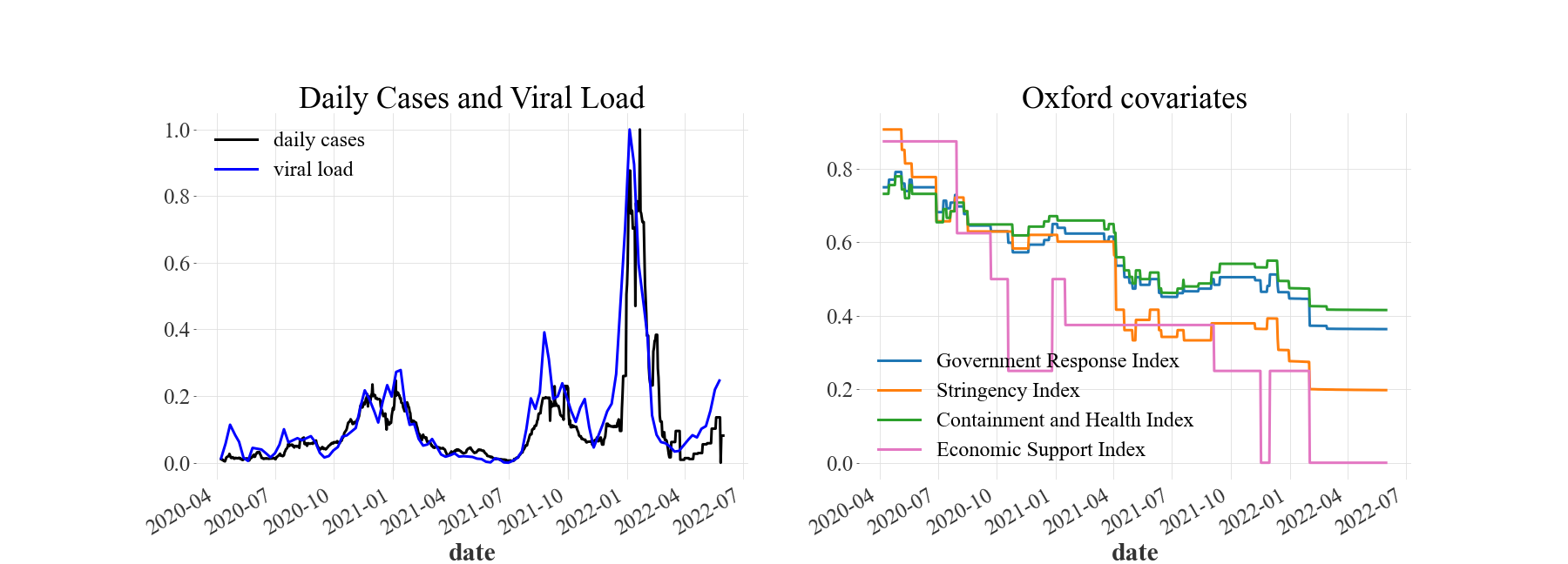}
         \caption{}
         \label{fig:Erie}
     \end{subfigure}
     \hfill
     \begin{subfigure}[b]{0.95\columnwidth}
         \centering
         \includegraphics[width=\textwidth]{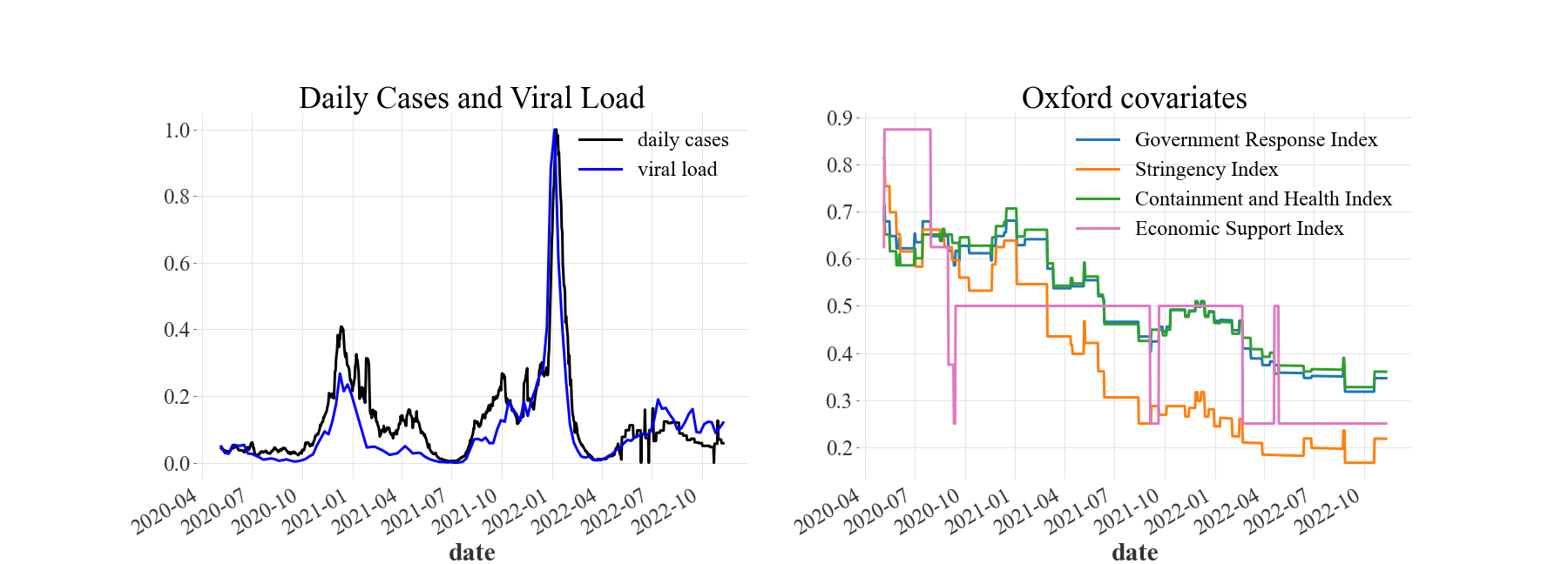}
         \caption{}
         \label{fig:Indiana}
     \end{subfigure}
        \caption{Processed data of two sample counties (a) Jefferson county, KY (b) Dauphin county, PA.}
        \label{fig:processed-data}
\end{figure}

Out of all counties in the Biobot dataset, we chose 13 counties with the most viral load measurements. In addition, two of the counties were kept as a holdout set to better assess the generalizability of our model by testing them on holdout counties. Fig. \ref{fig:processed-data} shows the processed time series of confirmed cases, viral load, and Oxford covariates that are input to our models.
We take 80\% of the historical data for each county as training and the rest as test data. Also, 10\% of the training data is taken as the validation data. We used validation data to identify the optimal hyperparameters of our models. We set the look-back window to 30 days and the forecast horizon to 10 days. For the implementation, we used the Darts package in Python~\cite{JMLR:v23:21-1177}, which is a comprehensive time series analysis package. Our codes can be found the project's GitHub page\footnote{github.com/mehrdadfazli/DeepLearning-COVID19-wastewater}

\subsection{Results and Discussion}

\begin{figure*}[!t]
    \centering
    \includegraphics[width=\textwidth]{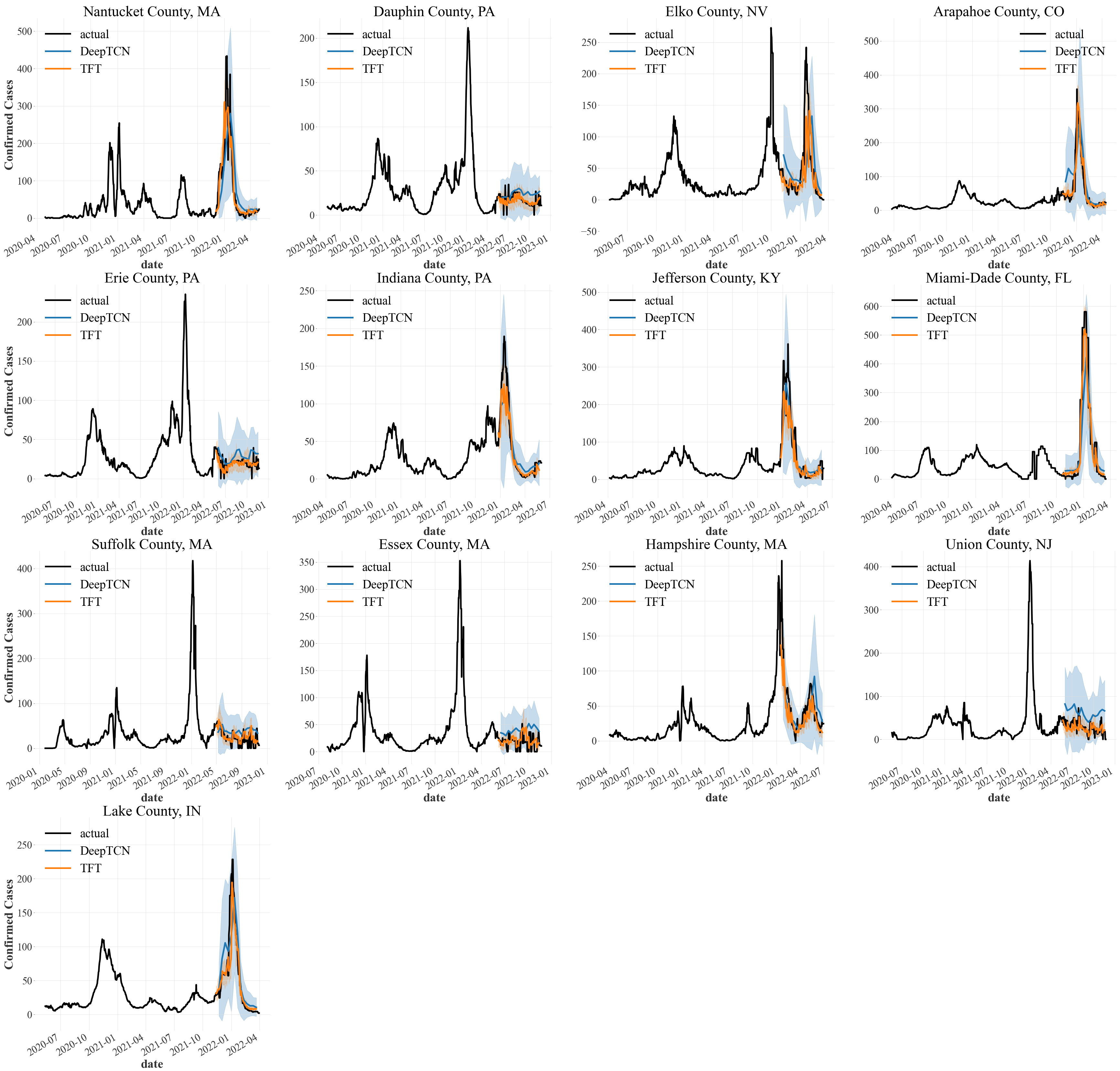}
    \centering
    \caption{Predictions of TFT and DeepTCN for 13 counties. Union county and Lake county are the two holdout counties.}
    \label{fig:All-preds-TFT-TCN}
\end{figure*}

We present the predictions of TFT and DeepTCN models for the 11 training counties and the two holdout counties (Fig.~\ref{fig:All-preds-TFT-TCN}). The predictions are the 10-day ahead point estimates and their corresponding 90\% confidence intervals associated with the test period for each county. DeepTCN's predictions are smoother but less accurate in some counties with considerably larger confidence intervals. On the other hand, TFT is relatively consistent and accurate for all counties with reasonably smaller confidence intervals. Another interesting observation is that the models, especially TFT, have effectively generalized to the two unseen counties. This can particularly be witnessed in the case of Lake county, as both models comfortably captured the spike in the confirmed cases in December 2021 and January 2022.

We use backtesting without retraining to make predictions and evaluate models. With backtesting, we shift the look-back window by a certain number of time steps and then make predictions for the forecast horizon. In our case, the forecast horizon and the stride (shift in the look-back window) are set to 10 days. Also, We have the choice to retrain the model before making predictions. In a real-world scenario, we retrain our model as new data becomes available and as frequently as the computational resources allow. However, retraining requires plenty of computational resources and many retraining iterations. For the sake of model comparison and evaluating the contribution of covariates, retraining is not necessary. Nonetheless, it is worth noting that by retraining the models with the availability of new data, one can achieve remarkably smaller errors in the predictions.

To facilitate the comparison of the models, we selected three metrics; mean absolute error (MAE), symmetric mean absolute percentage error (SMAPE), and coefficient of variation (CV) (equations \ref{eq:update2}-\ref{eq:update4}). MAE is the average raw difference between the predicted and actual time series. It is a simple yet interpretable and effective accuracy metric for forecasting tasks. We also chose SMAPE over MAPE since MAPE produces large errors when the actual series approaches zero. As our target series is daily confirmed cases, we have zeros in our target series, which prevents us from using MAPE. Instead, SMAPE is the symmetric version of the MAPE that can handle zeros in the target series. Also, similar to MAPE, SMAPE is a scale-free metric suitable for cross-studies comparisons. CV is another scale-free metric based on root mean square error (RMSE) expressed in percentage, which shows the variation in the errors with respect to the average of the actual series.

\begin{align}
    \label{eq:update2}
    & \operatorname{MAE} = \frac{1}{T}\Sigma_{t=1}^T|y_t - \hat{y}_t| \\
    & \operatorname{SMAPE} = 200\times\frac{1}{T}\Sigma_{t=1}^T \frac{y_t - \hat{y}_t}{|y_t| + \hat{y}_t} \\
    & \operatorname{CV} = 100\times\operatorname{RMSE}(y_t, \hat{y}_t)/\Bar{y_t}
    \label{eq:update4}
\end{align}

\begin{table}[!b]
\centering
\begin{subtable}{0.4\textwidth}
\centering
\caption{Training counties}\label{table:perf-train}
    \begin{tabular}{ c c c c }
     Model & MAE   & SMAPE & CV \\ \hline
     TFT                      & \textbf{16.00} & \textbf{41.20} & \textbf{75.88}\\ \hline
     TFT - no viral load      & 17.92 & 44.88  & 90.62\\ \hline
     DeepTCN                  & 25.95 & 59.30  & 100.80\\ \hline
     DeepTCN - no viral load  & 29.94 & 61.32  & 119.79\\ 
    \end{tabular}
\end{subtable}
\begin{subtable}{0.4\textwidth}
\centering
\caption{Holdout counties}\label{table:perf-holdout}
    \begin{tabular}{ c c c c }
     Model & MAE   & SMAPE & CV \\ \hline
     TFT                      & 10.73 & \textbf{32.92} & 56.10\\ \hline
     TFT - no viral load      & \textbf{9.92} & 36.50  & \textbf{42.88}\\ \hline
     DeepTCN                  & 20.98 & 51.17  & 83.35\\ \hline
     DeepTCN - no viral load  & 21.87 & 46.24  & 110.06\\ 
    \end{tabular}
\end{subtable}%
\vspace{12pt}
\caption{Performance comparison for the four models of TFT, TFT without viral load data, DeepTCN, and DeepTCN without viral load data.}\label{table:acc-tab}
\end{table}

To better validate the role of viral load data, we conducted an ablation study by training one TFT model and one DeepTCN model without the viral load data. Table~\ref{table:acc-tab} shows the performances of the four models for the three selected metrics. The first takeaway is that TFT outperforms DeepTCN by a considerable margin. We can also conclude that viral load has significantly helped the TFT and DeepTCN models produce more accurate predictions for training counties (Table \ref{table:perf-train}). Using viral load can reduce MAE by nearly $2$ units for TFT and $4$ units for DeepTCN. It also decreases SMAPE by more than 3\% for TFT and 2\% for DeepTCN.

Moreover, TFT allows for extracting the importance of the different variables. Although, this is different for the holdout counties as we only have two counties in the holdout set selected at random, and their data might have been easier to predict. On the other hand, the point of the holdout set is merely to assess the generalizability of our models, not to compare them. Comparing the results of table~\ref{table:perf-train} and table~\ref{table:perf-holdout}, one can verify that the models performed equally well for the unseen localities.

\begin{figure}
     \centering
     \begin{subfigure}{0.95\columnwidth}
         \centering
         \includegraphics[width=\textwidth]{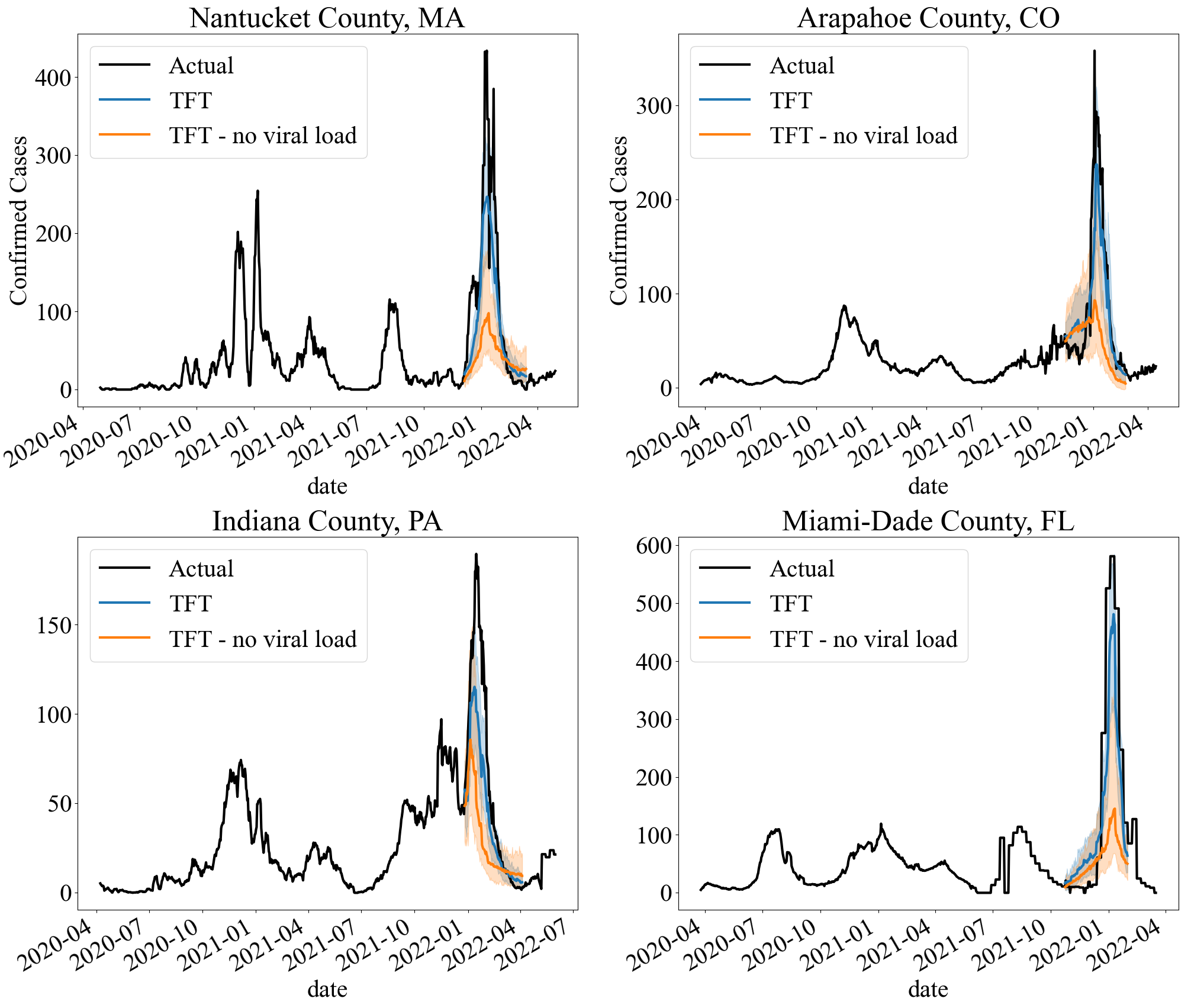}
         \caption{TFT models' predictions.}
         \label{fig:TFT-four-pred}
     \end{subfigure}
     \hfill
     \begin{subfigure}{0.95\columnwidth}
         \centering
         \includegraphics[width=\textwidth]{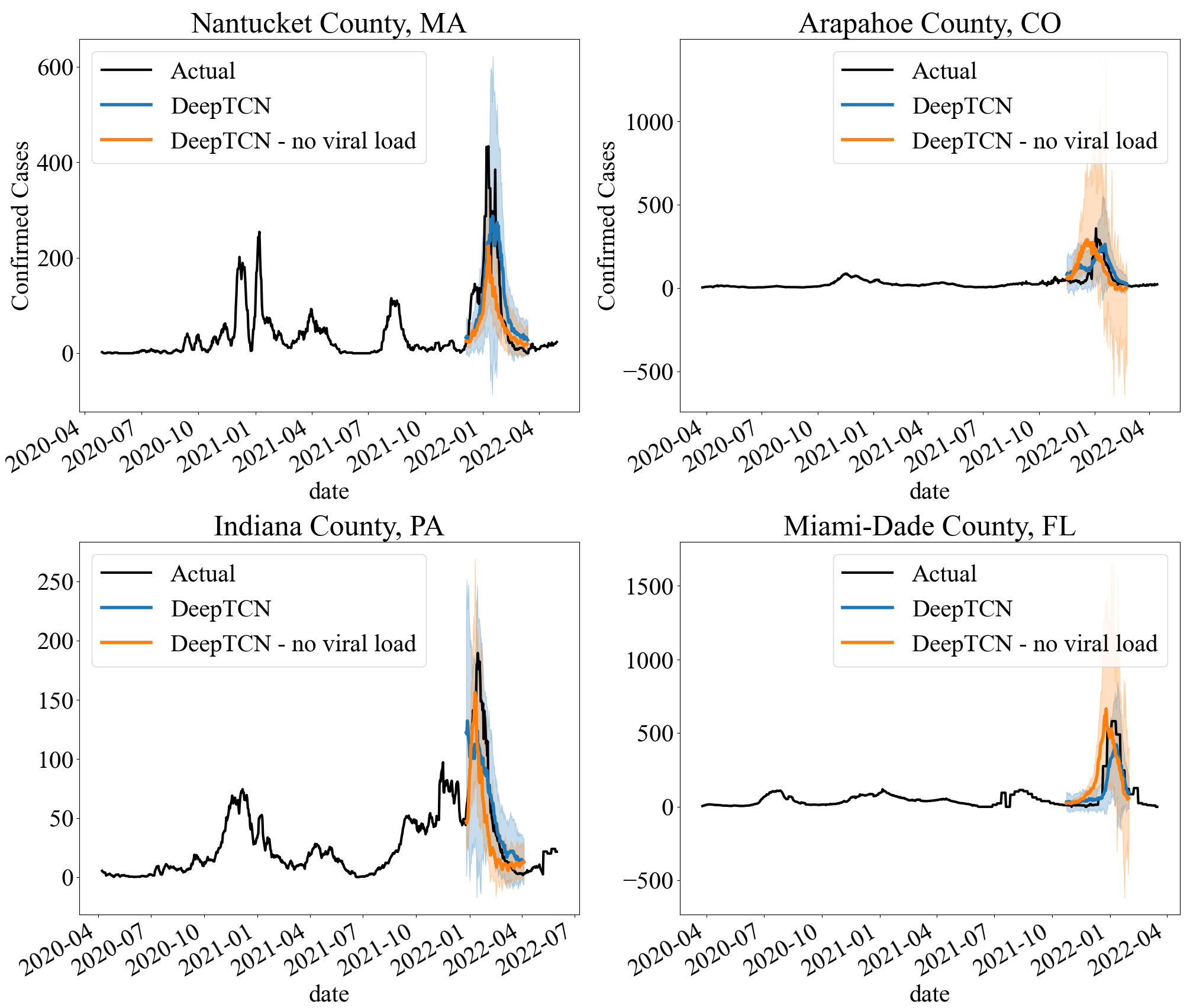}
         \caption{DeepTCN models' predictions.}
         \label{fig:DeepTCN-four-pred}
     \end{subfigure}
        \caption{Impact of viral load on (a) TFT and (b) DeepTCN models' predictions with 90\% confidence interval for four counties.}
        \label{fig:Pred-four}
\end{figure}

Fig.~\ref{fig:Pred-four} illustrates the information gained by the viral load data for TFT and TCN models by placing the predictions of each pair of models on the same plot for Nantucket county, Arapahoe county, Indiana county, and Miami-Dade county. Predictions are 10-day ahead forecasts that correspond to the test period of the time series. Compared with DeepTCN, the TFT model learns a better relationship between daily cases and viral load and successfully captures the abrupt rise in the number of cases in early 2022 (Fig.~\ref{fig:TFT-four-pred}). Moreover, supplementing viral load data to the models reduces the uncertainty in the model's predictions.

\begin{figure}
     \centering
     \begin{subfigure}{0.95\columnwidth}
         \centering
         \includegraphics[width=\textwidth]{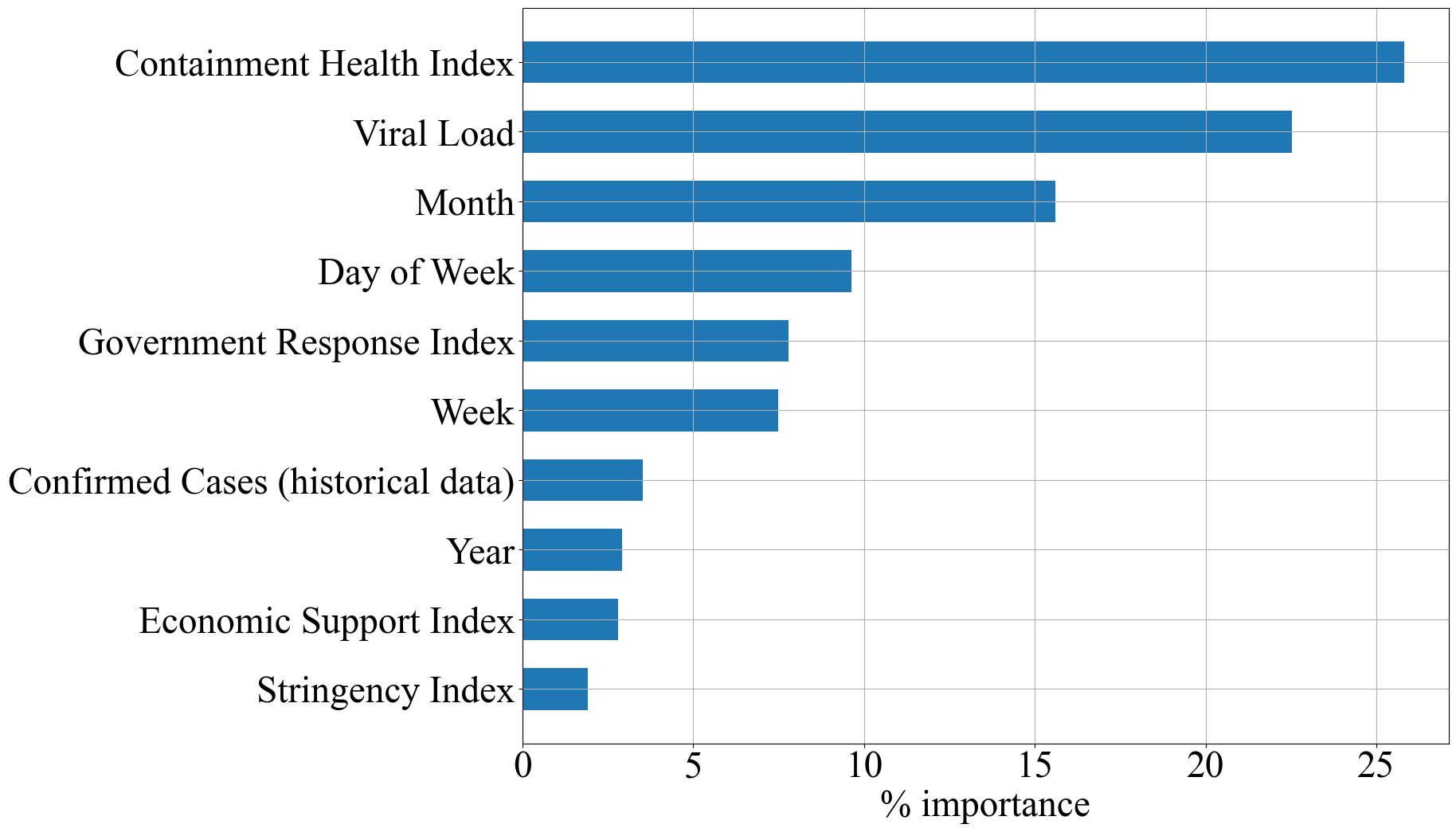}
         \caption{Encoder's variable importance.}
         \label{fig:imp-enc}
     \end{subfigure}
     \hfill
     \begin{subfigure}{0.95\columnwidth}
         \centering
         \includegraphics[width=\textwidth]{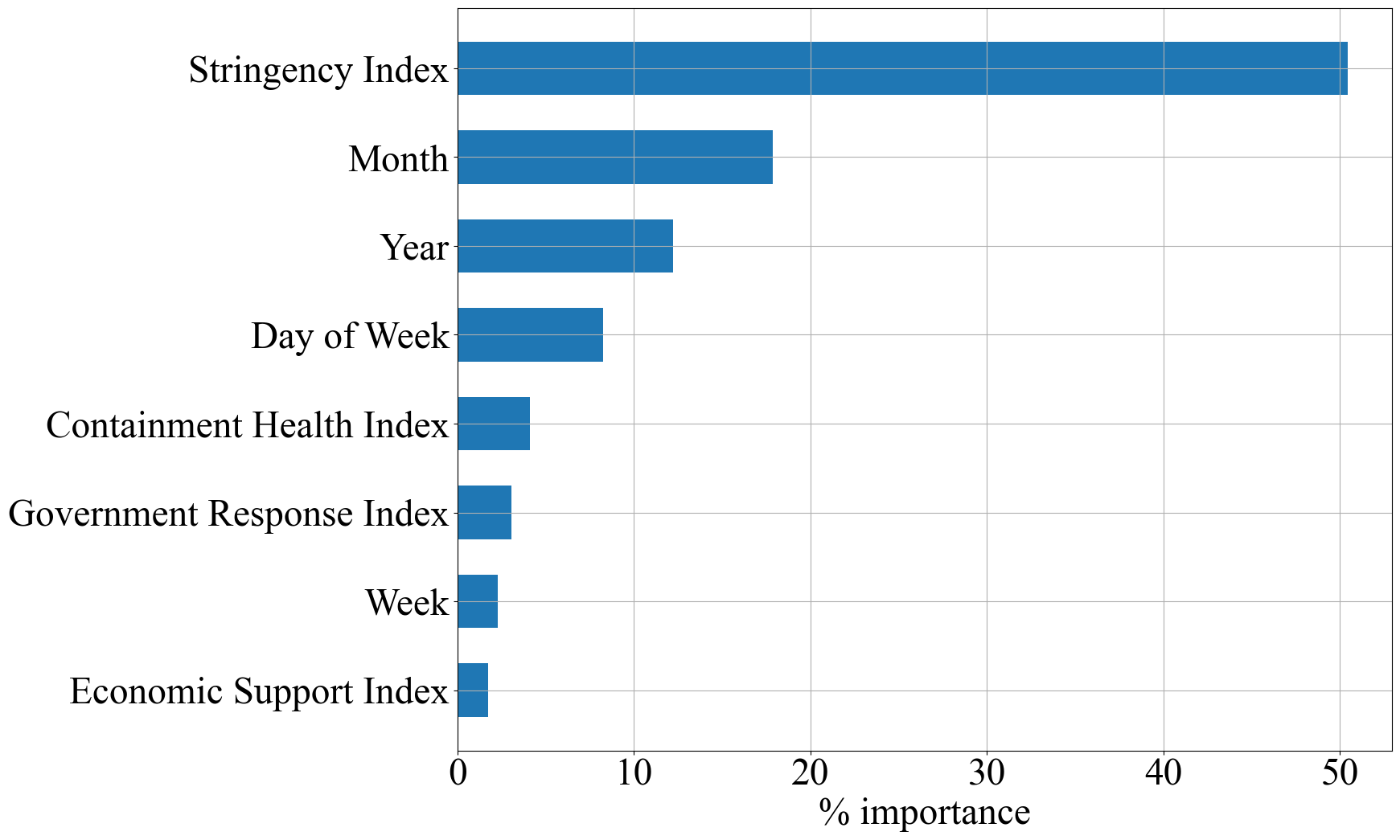}
         \caption{Decoder's variable importance.}
         \label{fig:imp-dec}
     \end{subfigure}
        \caption{Varibale importance for (a) encoder and (b) decoder.}
        \label{fig:imp-enc-dec}
\end{figure}

One of the significant advantages of TFT over other DL-based time series models is its transparency through its variable selection unit. Once the model is trained, we can aggregate the variable selection weights to measure how the model is attending to the input variables. As we have a variable selection module for the encoder and another for the decoder, we can obtain the variable importance separately. Fig.~\ref{fig:imp-enc} shows the percentage importance of the input variables of the encoder unit. The most significant input proves to be containment and health index. Containment accounts for school closings, public spaces closings, and other restrictions placed by local governments to slow down the spread of the virus~\cite{hale2021global}.
On the other hand, the health index measures healthcare access, quality, budget, and safety measures put in place by the officials, like contact tracing and public campaigning. The containment and health index is a weighted average of all the health and containment indices. The second important input is the viral load with over 20\% weight. This demonstrates the importance of viral load as a complementary data source for COVID-19 incidence forecasting. The next important inputs are the month and day of the week, which are associated with the seasonality in the data. This can be attributed to the seasonal pattern in social gatherings and superspreading events. 

On the other hand, the stringency index dominates the variable importance for the decoder (\ref{fig:imp-dec}) with ~50\% importance. The stringency index in OxCGRT is the containment and health index, excluding testing policy and contact tracing. Therefore, containment policies followed by the viral load are the most influential predictors in forecasting COVID-19 cases. It is noteworthy that the economic support index has the least importance out of all covariates considering both encoder and decoder variable importance.

\section{Conclusion}

In this study, we explored the possibility of leveraging deep learning to automatically learn the complex relationship between viral load and daily confirmed cases of COVID-19. Evidence has been abundant on the effectiveness of viral load in the early detection of an outbreak in communities. Nevertheless, there need to be more methodological approaches that incorporate viral load data in developing more accurate forecasting models. We proposed a deep learning framework to extract useful information in the viral load and enhance prediction accuracy. In addition to viral load, we augmented our data with socio-economic indicators and automatically generated seasonality covariates. We tested two multi-step probabilistic forecasting models that allow for covariates: DeepTCN and TFTFT. Our analysis showed the superiority of the TFT, a transformer-based model, over the DeepTCN model in extracting the relationship between viral load and COVID-19 incidence data. Despite DeepTCN, TFT offers some degree of transparency regarding how the model attends to the inputs. We demonstrated that containment policies and the viral load are essential factors in predicting daily cases of COVID-19.

This work can be extended using more data with finer granularity. Deep learning models unleash their full capacity when provided with a large amount of data. Also, the wastewater sampling frequency in Biobot's nationwide wastewater monitoring dataset is weekly. We interpolated viral load measurements for the days between two recordings which could introduce some errors in the analysis. Also, many counties implemented their wastewater surveillance system long after the CDC launched the national wastewater surveillance system. We dropped those counties in our analysis due to insufficient data. In addition to the viral load, daily cases provided with USA Facts also suffered from occasional misreports and corrections leading to inaccurate data for some counties. Moreover, with the growing data on wastewater viral load, adding other covariates, such as air quality index and temperature, could help the models learn a more holistic picture of the infection dynamics.

Wastewater signal is a rich source of pathogens that can monitor the development of infections in the communities. It has been used for over 40 years to track viral infections~\cite{sinclair2008pathogen}. COVID-19 underlined the necessity of establishing a robust wastewater surveillance system for better monitoring of an outbreak and taking timely preventative measures.

\bibliographystyle{IEEEtran}
\bibliography{ref.bib}

\end{document}